\documentclass{article}

\usepackage[utf8]{inputenc} %
\usepackage[T1]{fontenc}    %
\usepackage[bookmarks=true]{hyperref}
\usepackage{url}            %
\usepackage{booktabs}       %
\usepackage{amsfonts,amssymb,mathtools,amsthm}       %
\newtheorem{theorem}{Theorem}
\usepackage{nicefrac}       %
\usepackage{microtype}      %
\usepackage{cleveref}
\usepackage{siunitx}

\newcommand{\Lag}{\mathcal{L}} %

\newcommand\bbR{\ensuremath{\mathbb{R}}} %
\newcommand\bbE{\ensuremath{\mathbb{E}}} %
\newcommand\qdot{\ensuremath{\dot{q}}} %
\newcommand\qddot{\ensuremath{\ddot{q}}} %
\newcommand\M{\ensuremath{\mathbf{M}}} %

\newcommand\tq{\ensuremath{\tilde{q}}}
\newcommand\tqd{\ensuremath{\dot{\tilde{q}}}}
\newcommand\tqdd{\ensuremath{\ddot{\tilde{q}}}}

\usepackage{todonotes}
\usepackage{soul}
\definecolor{smoothgreen}{rgb}{0.7,1,0.7}
\sethlcolor{smoothgreen}

\usepackage[utf8]{inputenc}
\RequirePackage{luatex85}
\usepackage{pgfplots}
\pgfplotsset{compat=newest}
\pgfplotsset{every axis legend/.append style={%
cells={anchor=west}}
}
\usepgfplotslibrary{polar}
\usetikzlibrary{arrows}
\tikzset{>=stealth'}

\definecolor{C1}{rgb}{0.0, 0.447, 0.741}
\definecolor{C1_light}{rgb}{0.0, 0.6032388663967612, 1.0}
\definecolor{C2}{rgb}{0.85, 0.325, 0.098}
\definecolor{C3}{rgb}{0.929, 0.694, 0.125}
\definecolor{C4}{rgb}{0.494, 0.184, 0.556}
\definecolor{C5}{rgb}{0.466, 0.674, 0.188}
\definecolor{C6}{rgb}{0.301, 0.745, 0.933}
\definecolor{C7}{rgb}{0.635, 0.078, 0.184}

\definecolor{nice-red}{HTML}{E41A1C}
\definecolor{nice-orange}{HTML}{FF7F00}
\definecolor{nice-yellow}{HTML}{FFC020}
\definecolor{nice-green}{HTML}{4DAF4A}
\definecolor{nice-blue}{HTML}{377EB8}
\definecolor{nice-nice-red}{HTML}{984EA3}
\usepgfplotslibrary{groupplots}

\usetikzlibrary{shapes.geometric, arrows}

\tikzstyle{startstop} = [rectangle, rounded corners, minimum width=2cm, minimum height=1cm,text centered, draw=black, fill=none]
\tikzstyle{arrow} = [thick,->,>=stealth]

\usetikzlibrary{backgrounds}

\usepackage{ifthen}

\usepackage{arxiv}

\usepackage[numbers]{natbib}
\usepackage{multicol}
\usepackage{balance}
\usepackage[utf8]{inputenc} %
\usepackage[T1]{fontenc}    %
\usepackage{hyperref}       %
\usepackage{url}            %
\usepackage{booktabs}       %
\usepackage{amsfonts}       %
\usepackage{nicefrac}       %
\usepackage{microtype}      %
\usepackage{lipsum}		%
\usepackage{graphicx}

\title{Training Structured Mechanical Models by Minimizing Discrete Euler-Lagrange Residuals}
\renewcommand{\shorttitle}{Training SMMs by Minimizing DEL Residuals}
\renewcommand{\undertitle}{}
\renewcommand{\headeright}{}

\author{Kunal Menda\\
Aeronautics \& Astronautics\\
Stanford University\\
Stanford, CA 94305\\
\texttt{kmenda@stanford.edu}\\
\And
Jayesh K. Gupta\\
Autonomous Systems\\
Microsoft\\
Redmond, WA 98052\\
\texttt{jayesh.gupta@microsoft.com}\\
\AND
Zachary Manchester\\
Robotics Institute\\
Carnegie Mellon Univeristy\\
Pittsburgh, PA 15213\\
\texttt{zmanches@andrew.cmu.edu}\\
\And
Mykel J. Kochenderfer\\
Aeronautics \& Astronautics\\
Stanford University\\
Stanford, CA 94305\\
\texttt{mykel@stanford.edu}
}

\date{}

\begin{document}

\maketitle

\begin{abstract}
Model-based paradigms for decision-making and control are becoming ubiquitous in robotics. 
They rely on the ability to efficiently learn a model of the system from data. 
Structured Mechanical Models (SMMs) are a data-efficient black-box parameterization of mechanical systems, typically fit to data by minimizing the error between predicted and observed accelerations or next states. 
In this work, we propose a methodology for fitting SMMs to data by minimizing the discrete Euler-Lagrange residual.
To study our methodology, we fit models to joint-angle time-series from undamped and damped double-pendulums, studying the quality of learned models fit to data with and without observation noise.
Experiments show that our methodology learns models that are better in accuracy to those of the conventional schemes for fitting SMMs. 
We identify use cases in which our method is a more appropriate methodology.
\end{abstract}

\section{Introduction}
\label{sec:introduction}

Many autonomous systems implement model-based control schemes, in which the system uses an internal model of itself, its environment, and the consequences of its own actions.
In robotic settings, these models typically map the robot's configuration, its velocities, and the forces applied to it to the acceleration on the robot's degrees-of-freedom.
There is a growing need to efficiently learn these models from potentially noisy time-series describing the evolution of the system.
Structured Mechanical Models (SMMs)~\cite{Gupta2020, gupta2019general} provide a black-box but data-efficient parameterization of mechanical systems that have been shown to be well-suited to the task of learning models of dynamics from data.

Instead of directly parameterizing a function that predicts accelerations given the robot configuration, velocity, and inputs, SMMs parameterize the Lagrangian of a mechanical system.
The accelerations on the system are then computed from the Lagrangian using the Euler-Lagrange equation~\cite{Gupta2020,gupta2019general, lutter2018deep, cranmer2020lagrangian}.
Prior work fit SMMs by minimizing the error between the predicted and observed accelerations~\cite{lutter2018deep, cranmer2020lagrangian}, or the predicted and observed next-states of the system~\cite{Gupta2020}.
In cases in which dynamics are continuous, time-series of observations can typically be filtered or smoothed in order to recover estimates of the system's configurations, velocities, and accelerations, from which we can train a model. 
However, in the case where dynamics are inherently discontinuous, such as domains involving contact, a continuous formulation of Euler-Lagrange equations will yield predictions of infinite acceleration, and is thus inappropriate.

In this work, we propose an alternative methodology for fitting SMMs to a time-series of system configuration measurements. 
We observe that for a trajectory to have been generated by a given Lagrangian, the discrete Euler-Lagrange (DEL) equations must be zero along the trajectory \cite{marsden2001discrete}.
We therefore propose to minimize the DEL \textit{residual}, attempting to find a Lagrangian such that the DEL equations are zero along the trajectories being fit.
This approach is not sufficient in itself, however, due to \textit{gauge invariance}.
Specifically, if the DEL equations are satisfied along a trajectory given a Lagrangian $\Lag$, 
then they will also be satisfied given a Lagrangian $\alpha \Lag + \beta$, for $(\alpha, \beta) \in \bbR$.
As a result, a Lagrangian that is everywhere equal to a constant satisfies the DEL equations.
To avoid this degenerate solution, we propose a regularization term that ensures the learned Lagrangian is non-constant along the trajectory.

To validate our approach, we fit SMMs to data from damped and undamped double-pendulums.
We show that when training with our method instead of using acceleration or next-state regression, we are able to reliably learn models with lower error.
We demonstrate that this result holds true when fitting SMMs to data recovered from noisy observations of an undamped and damped double-pendulum.

The contributions of this work are as follows:
\begin{itemize}
    \item We propose an optimization objective for fitting SMMs based on the discrete Euler-Lagrange residual, 
    \item We propose a regularization term that guarantees our method does not find degenerate solutions, and,
    \item We demonstrate that SMMs fit with our method are of better quality than those fit with conventional approaches.
\end{itemize}

We conclude this work with a discussion of potential application domains in which minimizing the DEL-residual is more appropriate than acceleration or next-state regression, such as domains involving contact.
The codebase containing the experiments conducted and an example implementation of our methodology can be found at \url{https://github.com/sisl/delsmm}.

\section{Background}
\label{sec:background}

\subsection{Lagrangian Dynamics}
Consider a configuration space of a mechanical system $q\in \bbR^n$ and associated velocity $\qdot \in \bbR^n$. 
The evolution of the system can be understood by specifying the \textit{Lagrangian} of the system, i.e.
\begin{equation}
    \Lag(q,\qdot) %
        = \frac12 \qdot^\top \M(q) \qdot - V(q)
\end{equation}
where $\frac12 \qdot^\top \M(q) \qdot$ and $V(q)$ are the kinetic and potential energy of the system, respectively, and $\M(q) \succ 0$ is the \textit{mass-matrix}.
The system maybe also be \textit{forced} via an external forcing function $F(q,\qdot, u)$, where $u$ is a control-input. 
Note that these forces are \textit{non-conservative} in that they change the total-energy of the system.
For simplicity, we considered uncontrolled systems in this work, and thus drop the dependence on $u$.

\subsubsection{The Euler-Lagrange Equation}
The Lagrangian of a system can be used to specify the dynamics of the system via the \textit{Euler-Lagrange} equation.
The continuous-time version of this equation is:
\begin{equation}
    \frac{d}{dt}\left(\frac{\partial \Lag}{\partial \qdot}\right) = \frac{\partial \Lag}{\partial q} + F(q,\qdot)
\end{equation}
This equation can be used to find the accelerations $\qddot$ on the system, as follows:
\begin{equation}
\label{eqn:acc}
    \qddot = \left(\frac{\partial^2 \Lag}{\partial \qdot^2}\right)^{-1}\left[ F(q,\qdot) + \frac{\partial \Lag}{\partial q} - \left(\frac{\partial^2 \Lag}{\partial \qdot\partial q}\right)\qdot  \right]
\end{equation}
The trajectory of the system given $q_0, \qdot_0$ can then be found by a numerical integration scheme such as Runge-Kutta~\cite{runge1895}.

Suppose we instead are given $q_0, q_1$, which are the configurations of the system at the first and second time-step of interest. 
We can find the configuration $q_3$ that succeeds these initial conditions by considering a \textit{discrete} formulation of the Lagrangian and the Euler-Lagrange equation~\cite{marsden2001discrete}.
We define the discrete-Lagrangian $\Lag_d$ and discrete generalized-force $F_d$ given simulation step-size $h$ to be:
\begin{equation}
\begin{aligned}
    \Lag_d(q_1, q_2, h) &= h\Lag\left(\frac{q_1 + q_2}{2}, \frac{q_2-q_1}{h}\right)\\
    F_d(q_1,q_2,h) &= h F\left(\frac{q_1 + q_2}{2}, \frac{q_2-q_1}{h}\right)
\end{aligned}
\end{equation}
and the discrete Euler-Lagrange (DEL) equation as:
\begin{equation}
\label{eqn:deleqn}
\begin{aligned}
    \text{DEL}(q_{1:3}) &= D_2 \Lag_d(q_1, q_2) + D_1 \Lag_d(q_2,q_3) \\
    &\quad + \frac{1}{2}\left( F_d(q_1,q_2) + F_d(q_2, q_3) \right)\\
    &= 0
\end{aligned}
\end{equation}
Here we use the \emph{slot derivative} $D_i$ to indicate partial differentiation with respect to a function's $i$-th argument.

\subsubsection{Gauge Invariance}
It should be noted that, given a system's dynamics, the corresponding Lagrangian is not unique.
Specifically, if $\tilde{\Lag} = \alpha \Lag + \beta$ for some $(\alpha, \beta) \in \bbR$, then it is trivial to see from \Cref{eqn:acc} and \Cref{eqn:deleqn} that the dynamics that correspond to $\Lag$ and $\tilde{\Lag}$ are identical.
That is, if there is a tuple $(q, \qdot, \qddot)$ and Lagrangian $\Lag$ for which \Cref{eqn:acc} holds, or a tuple $q_{1:3}$ and discrete Lagrangian $\Lag_d$ for which \Cref{eqn:deleqn} holds, then it will also hold for $\tilde{\Lag} = \alpha \Lag + \beta$.

\subsection{Structured Mechanical Models}
Many works in recent years have proposed a black-box parameterization for the dynamics of a mechanical system~\cite{gupta2019general, Gupta2020, lutter2018deep, cranmer2020lagrangian, greydanus2019hamiltonian}. 
\citet{gupta2019general} and \citet{lutter2018deep} proposed to parameterize the components of the Lagrangian of a dynamical system using neural networks, and then derive the accelerations on the system via \Cref{eqn:acc}. 
Specifically, they proposed to parameterize the Cholesky-factor of $\M(q)$ using a neural-network mapping $q\in\bbR^n\rightarrow \bbR^{\frac{n^2-n}{2}}$, and another neural network representing the potential-energy $V(q)$ of the system, mapping $q\in\bbR^n\rightarrow \bbR$.
\citet{gupta2019general} also propose to parameterize the generalized forces $F(q,\qdot,u)$ using a neural network, and show that the expressive power of this model is equivalent to that of a neural network directly mapping $(q,\qdot,u)$ to $\qddot$, though with considerably better generalization properties. 
\citet{cranmer2020lagrangian} extend these works by parameterizing the Lagrangian as any neural-network mapping $(q,\qdot)\rightarrow \bbR$, allowing for the use of novel architectures such as graph neural-networks. 

Each of these works refers to these parameterizations by a different name, such as Deep Lagrangian Networks~\cite{lutter2018deep}, and Lagrangian Neural Networks~\cite{cranmer2020lagrangian}. 
We follow the naming convention of \citet{Gupta2020}, calling a model parameterizing $\M(q), V(q)$ and $F(q,\qdot)$ with neural networks a \textit{structured mechanical model} (SMM).

\subsubsection{Parameter Optimization in Prior Work}

We are given a time-series of sampled system configurations $y_{1:T}$, possibly corrupted by some observation noise.
It is common to use a \textit{smoothing} routine to try to recover a time-series of the system's configurations $\tq$, velocities $\tqd$, and accelerations $\tqdd$ from $y_{1:T}$.
One popular approach that is used by this paper is to assume the system is a double-integrator and find the desired series using Kalman smoothing \cite{aravkin2017generalized}.

The parameters of an SMM are thus optimized to minimize the empirical risk between $\tqdd$ and the acceleration predicted by the SMM given $\tq$ and $\tqd$~\cite{lutter2018deep, cranmer2020lagrangian}.
\citet{Gupta2020} propose instead to minimize the empirical risk between the predicted $\tq', \tqd'$ and smoothed next-states given $\tq, \tqd$.
We refer to these two approaches as \textit{acceleration} and \textit{next-state regressions} respectively.

In the next section, we propose an alternative methodology for optimizing the parameters of an SMM.
\section{Methodology}

As shown in \Cref{sec:background}, a system's dynamics can be derived from its Lagrangian using a continuous or discrete-time formulation. 
In this section, we describe a methodology for fitting an SMM to data using the discrete Euler-Lagrange equation. 
Again, we assume we have access to a time-series $\tq_{1:T}$ found by smoothing an observation time-series, but do not assume access to the time-derivatives of this series.

We begin by observing that, given an arbitrary tuple of configurations $q_{1:3}$, we do not expect $\text{DEL}(q_{1:3}) = 0$ as in \Cref{eqn:deleqn}.
We define the \textit{DEL residual} $\rho(q_{1:3})$ to be:
\begin{equation}
    \rho(q_{1:3}) = \lVert \text{DEL}(q_{1:3}) \rVert_2^2
\end{equation}
The essence of our methodology is to find the parameters $\theta$ of an SMM by minimizing the DEL-residual.

Unfortunately, due to gauge-invariance, a trivial solution exists that has a zero DEL residual while incorrectly estimating the dynamics. 
Specifically, this solution is a constant Lagrangian, i.e. $\Lag(q,\qdot) = \beta, F(q,\qdot) = 0~\forall (q,\qdot)$, for any constant $\beta \in \bbR$, which gives a zero DEL residual for all trajectories.
We can avoid this trivial solution by adding the constraint that the mass-matrix is strictly \emph{positive definite}. 
We incorporate this constraint by adding a barrier-function commonly used in semidefinite programming that allows us to lower-bound the minimum eigenvalue of the mass-matrix.
Let $\rho(q_{1:3} \mid \theta)$ be the DEL-residual given that we parameterize the SMM components $M_\theta$, $V_\theta$, and $F_\theta$.
The loss $L(\theta)$ is defined as follows:
\begin{equation}
\label{eqn:delloss}
\begin{aligned}
    L(\theta) =& \overset{L_1(\theta)}{\overbrace{\bbE_{\tilde{q}_{1:3}\sim \mathcal{D}}\left[ \rho(\tilde{q}_{1:3} \mid \theta)\right]}} \\
    &- \mu \underset{L_2(\theta)}{\underbrace{\bbE_{\tilde{q}\sim \mathcal{D}}\left[\log\det(\M_\theta(\tilde{q}) - \alpha I)\right]}}
\end{aligned}
\end{equation}
where $\alpha$ is an arbitrarily chosen eigenvalue lower-bound, and $\mathcal{D}$ is the training dataset.
The addition of the regularizer ensures that $\min \text{eig}(M_\theta(\tilde{q})) > \alpha~\forall~\tilde{q}\in \mathcal{D}$.
A sensible choice for $\alpha$ is simply a fraction smaller than the smallest eigenvalue of the mass-matrix over the dataset given the initial parameter guess $\theta_0$.
We show in \Cref{thm:unbiased} that minimizing this loss as opposed to the DEL-residual does not yield a biased solution.

\begin{theorem}
\label{thm:unbiased}
Let the model class $\mathcal{M} = \{\M_\theta, V_\theta, F_\theta\}$ be closed under multiplication by a positive scalar. 
Then, there exists a minimizer $\theta$ of $L(\theta)$ that is also a non-trivial minimizer of $L_1(\theta)$.
\end{theorem}

\def\tg{\tilde{\theta}^*_\gamma}
\def\ddt{\frac{d}{d\theta}}
\begin{proof}
If $\mathcal{M} = \{\M_\theta, V_\theta, F_\theta\}$ is closed under multiplication by a positive scalar, then there exists $\tilde{\theta}_\gamma(\theta, \gamma)$ such that:
\begin{equation}
\begin{aligned}
\forall~(q,\qdot)~&(\gamma \M_\theta(q) = \M_{\tilde{\theta}_\gamma}(q)) \\ \wedge&(\gamma V_\theta(q) = V_{\tilde{\theta}_\gamma}(q)) \\ \wedge& (\gamma F_\theta(q,\qdot) = F_{\tilde{\theta}_\gamma}(q,\qdot)).
\end{aligned}
\end{equation}
Furthermore, let:
\begin{equation}
\begin{aligned}
    \theta^* = ~&\underset{\theta}{\arg\min}\quad &L_1(\theta) \\
    & \textrm{s.t.} \quad & \M_\theta(q) \succ 0~\forall~q\in \mathcal{D}
\end{aligned}
\end{equation}
be a non-trivial minimizer of $L_1(\theta)$, and let $\tg = \tilde{\theta}_\gamma(\theta^*, \gamma)$. 
Because differentiation is linear, we know that:
\begin{equation}
\begin{aligned}
    \ddt L(\tg) &= \ddt L_1(\tg) - \mu \ddt L_2(\tg). \\
\end{aligned}
\end{equation}
By gauge-invariance, 
\begin{equation}
\begin{aligned}
    \ddt L_1(\tg) &= 0 \\
\end{aligned}
\end{equation}
and as $\gamma \rightarrow \infty$:
\begin{equation}
\begin{aligned}
    \ddt L_2(\tg) &= \ddt \bbE_{\tilde{q}\sim \mathcal{D}}\left[\log\det(\M_{\tg}(\tilde{q}) - \alpha I)\right]
    \\
    &\approx \ddt \bbE_{\tilde{q}\sim \mathcal{D}}\left[\log\det(\M_{\tg}(\tilde{q}))\right] \\
    &= \frac{1}{\gamma}\ddt \bbE_{\tilde{q}\sim \mathcal{D}}\left[\log\det(\M_{\theta^*}(\tilde{q}))\right]\\
    &\rightarrow 0.
\end{aligned}
\end{equation}
Therefore, as $\gamma \rightarrow \infty$,
\begin{equation}
    \ddt L(\tg) \rightarrow 0
\end{equation}
implying that there is exists a minimizer of $L(\theta)$, i.e. $\tilde{\theta}^*_{\gamma\rightarrow\infty}$, that is also a minimizer of $L_1(\theta)$.
\end{proof}

In the next section, we compare the accuracy of SMMs learned using acceleration and next-state regression to those learned by minimizing \Cref{eqn:delloss}.
\section{Experiments}

\begin{figure}[t]
    \centering
    \includegraphics[width=0.6\columnwidth]{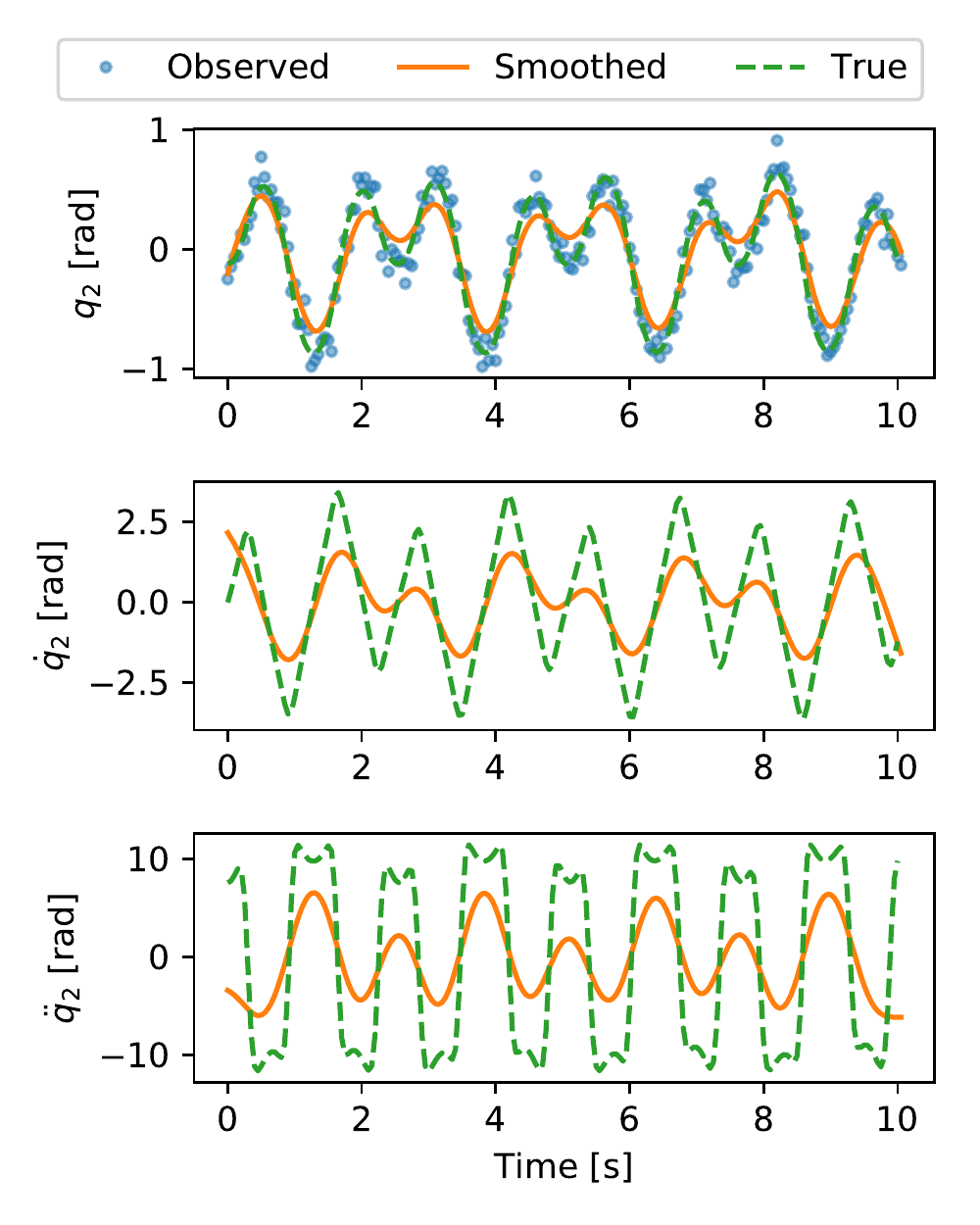}
    \caption{Observed, true, and smoothed trajectories used for training.}
    \label{fig:smoothed}
\end{figure}

\begin{figure}[t]
    \centering
    \includegraphics[width=0.45\columnwidth]{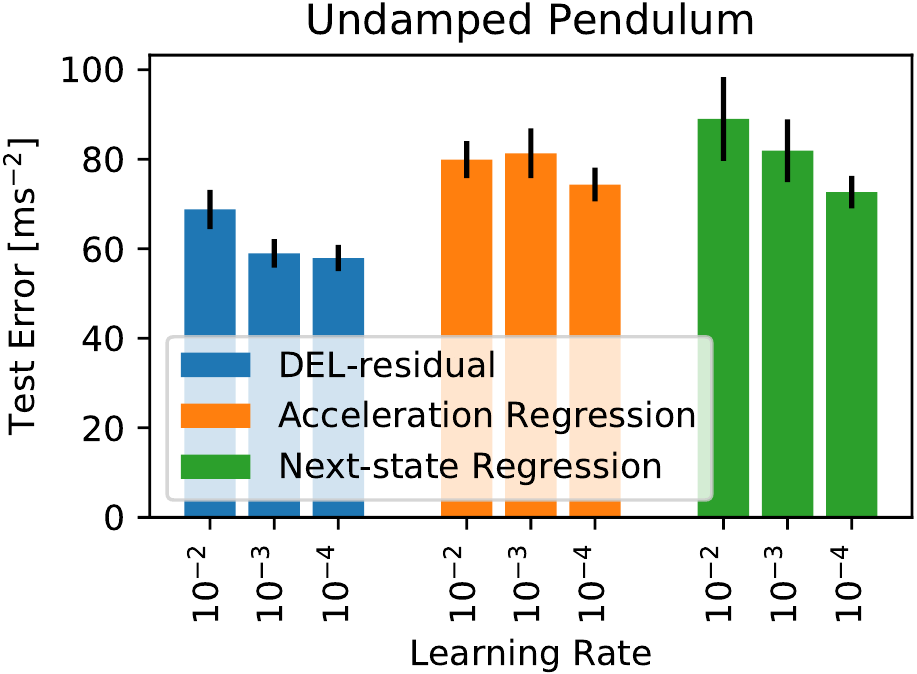}
    \includegraphics[width=0.465\columnwidth]{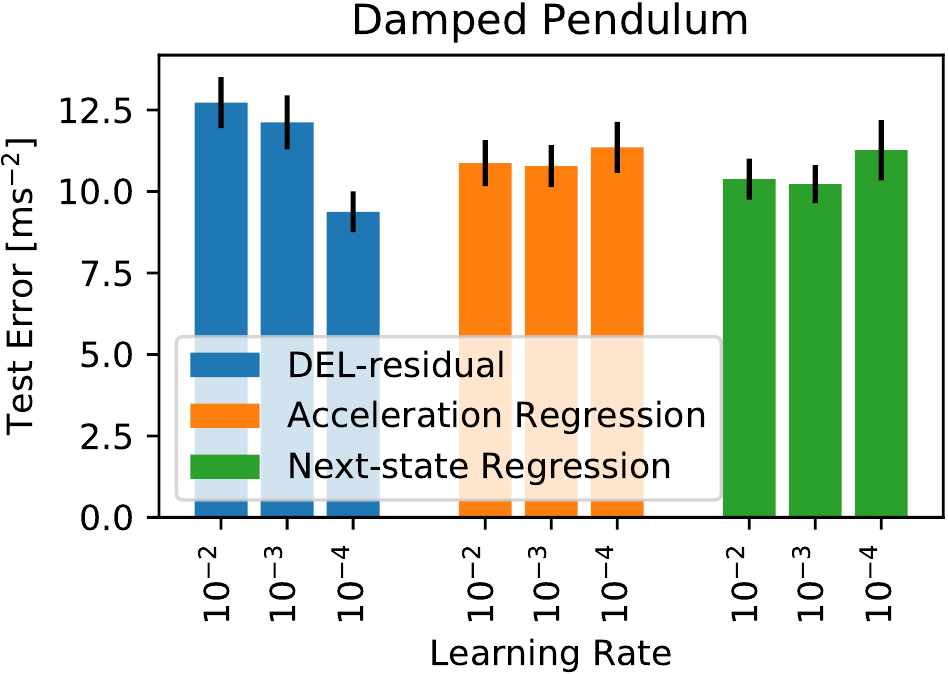}
    \caption{Comparison of training methodologies on data from a damped and undamped double pendulum.}
    \label{fig:undamped}
\end{figure}

\label{sec:experiments}
This section demonstrates that, when data is collected in the presence of noise, the SMMs learned by minimizing \Cref{eqn:delloss} are more accurate than those learned by acceleration or next-state regression.
We do so by studying a damped and undamped double-pendulum for which we learn $F_\theta$ and set it to zero, respectively.

\subsection{Double Pendulum Domain}
We simulate a double pendulum by specifying its Lagrangian and the forces that act on it.
The mass matrix and potential energy forming the Lagrangian are specified in \Cref{app:dpendyn}. 
In the undamped setting, the forces $F(q,\qdot)$ that act on the system are zero, and in the damped case, the forces are $F(q, \qdot) = -\eta \circ \qdot$, for some positive $\eta \in \bbR^2$, where $\circ$ represents the Hadamard product. 

The double pendulum is simulated for 200 time-steps using a step-size of 0.05 and a variational integration scheme. 
We initialize the system at rest with joint-angles uniformly distributed between $[-\pi/2, \pi/2)$. 
We simulate 16 trajectories, using 8 trajectories as part of the training set, 4 as part of a validation set, and 4 as part of a test set. 
The trajectories comprising these sets are randomized across seeds.
We observe the joint angles of systems at every time-step with an additive Gaussian observation noise with zero-mean and a standard deviation of 0.1 $\si\radian$, yielding observations $y_{1:T}$.
We also conduct the experiment using noise standard deviations of 0.05 and 0.4 $\si\radian$ and present the results in \Cref{app:noiseexp}.

\subsection{Experimental Protocol}

We estimate the smoothed joint angles $\tq_{1:T}$, velocities $\tqd_{1:T}$, and accelerations $\tqdd_{1:T}$ from $y_{1:T}$.
Specifically, we assume the data is generated from a linear dynamical system with dynamics:
\begin{equation}
    \begin{aligned}
    x_{t+1} = \begin{bmatrix} q_{t+1}\\\qdot_{t+1}\\\qddot_{t+1} \end{bmatrix} &= \exp\left(\begin{bmatrix}0 & 1 & 0\\0 & 0 & 1\\0&0&0\end{bmatrix}\Delta t\right)x_t  + w_t\\ 
    y_t &= \begin{bmatrix}1&0&0\end{bmatrix}~x_t +v_t
    \end{aligned}
\end{equation}
where $w_t \sim \mathcal{N}( 0, Q)$ and $v_t  \sim \mathcal{N}(0, R)$. 
We use expectation-maximization to find the likelihood-maximizing observation covariance matrix and the distributions over initial conditions. 
We use a fixed covariance matrix of $Q=\text{diag}([\num{e-3},\num{e-3},1.0])$.
A sample trajectory recovered from this smoothing procedure is depicted in \Cref{fig:smoothed}, which shows that high-frequency content is not recovered well in the smoothed accelerations.
 
We fit to the smooth data by minimizing the appropriate loss using stochastic gradient descent.
Specifically, we use a batch size of $256$ tuples, and use Adam~\cite{kingma2014adam} to optimize with varied initial learning rates $\xi_0$.
All learning rates follow a decay schedule of $\xi_k = 500 \xi_0 /(500 + k) $, where $k$ is the training epoch.
We minimize each loss for 500 epochs.

To avoid overfitting, we select a model with the lowest predicted acceleration error on the validation set. 
When selecting this model, we use smoothed validation data.
We then evaluate the quality of the fit by comparing the predicted and true accelerations on the test data.
The model with the lowest error in predicted acceleration is deemed the best performing model.

When fitting to data from an undamped double pendulum, we use a conservative SMM in which $F_\theta = 0$.
When fitting to data from a damped double pendulum, we represent $F_\theta$ with a neural network taking inputs $q$ and $\qdot$.

To compare the three methodologies (minimizing the regularized DEL-residual, acceleration regression, and next-state regression), we randomize the trajectories in the train, test, and validation datasets, as well as the initialization of the SMM, over 10 random seeds.
We report the mean error in predicted accelerations on the test set, as well as the standard error of this estimate.
For each methodology, we compare results for learning rates $\xi_0 \in \{ \num{e-2}, \num{e-3}, \num{e-5} \}$.

\subsection{Results}
In \Cref{fig:undamped}, we present the mean and standard errors of test performance for various learning rates, and for all methodologies.
We see that in both cases, the quality of models learned by minimizing the DEL-residual via \Cref{eqn:delloss} yields improvements over using either of the baseline methodologies, for an appropriate choice of learning rate.
We present additional experiments for different noise standard deviations in \Cref{app:noiseexp}, which support these conclusions.

The experiments conducted suggest that fitting SMMs to time-series by minimizing the DEL-residual via \Cref{eqn:delloss} yields models that are better than those learned by fitting SMMs via acceleration or next-state regression.
\section{Conclusion}
\label{sec:conclusion}

In this work, we proposed a methodology for fitting SMMs to a time-seires of observed states by minimizing the discrete Euler-Lagrange residual. 
To avoid allowing a trivial solution to be found, we introduced a regularization term that guarantees that the mass matrix has a lower-bounded eigenvalue for all states in the dataset.
We proved that using the regularized loss does not bias the learning objective.
Furthermore, we showed in experiments on nosiy data from damped and undamped pendulums that our methodology learns better quality solutions than acceleration or next-state regression.

An application of particular interest is the fitting of SMMs to contact-rich systems.
In such systems, the use of the continuous formulation of the Euler-Lagrange equations is inappropriate, owing to the presence of infinitely large accelerations on contact events.
The discrete formulation of these equations straightforwardly incorporates contact by introducing inequality constraints while minimizing the DEL-residual. 
Extending our methodology to enable the fitting of SMMs to contact-rich avenues is a promising avenue for future work.

\medskip
\bibliographystyle{plainnat}
\bibliography{references}

\onecolumn
\appendix
\section{Experiments}
\label{app:exp}
In this section, we detail the dynamics of the double pendulum used in our experiments, and display the result of experiments performed with additional noise standard deviations.
\subsection{Double Pendulum Dynamics}
\label{app:dpendyn}
We use double pendulum dynamics in our experiments.
The mass matrix $\M_{sys}(q)$ and potential energy $V_{sys}(q)$ of the double pendulum are as follows: 
\begin{equation}
    \M_{sys}(q) = \begin{bmatrix} 
        I_{11} & I_{12} \\
        I_{12} & I_2
    \end{bmatrix}
\end{equation}
where
\begin{align}
    I_1 &= \frac{1}{3} m_1 l_1^2 \\
    I_2 &= \frac{1}{3} m_2 l_2^2 \\
    I_{11} &= I_1 + I_2 + m_2 {l_1}^2 + m_2 l_1 l_2 \cos{q_2} \\
    I_{12} &= I_2 + \frac{1}{2} m_2 l_1 l_2 \cos{q_2}
\end{align}
\begin{equation}
    V_{sys}(q) = - \frac{1}{2} m_1 g l_1 \cos{q_1} - m_2g\left(l_1 \cos{q_1} + \frac{l_2}{2}\cos(q_1+q_2)\right)
\end{equation}

For the experiments, we used the parameters specified in \Cref{tab:sysparams}.
\begin{table}[h!]
\centering 
\caption{System dynamics parameters.}
\begin{tabular}{ccs|ccs} \toprule
    {Parameter} & {Value} & {Unit}&{Parameter} & {Value} & {Unit} \\ \midrule
    $m_1$  & 1.0 & \si{\kilogram} & $\eta_1$  & 0.5 & \si{\newton\meter\second\per\radian} \\  
    $m_2$  & 1.0 & \si{\kilogram} & $\eta_1$  & 0.5 & \si{\newton\meter\second\per\radian} \\
    $l_1$  & 1.0 & \si{\meter} & & & \\
    $l_2$  & 1.0 & \si{\meter} & & & \\
    $g$  & 10.0 & \si{\meter\per\second^2} & & & \\ \bottomrule
\end{tabular}
\label{tab:sysparams}
\end{table}

\subsection{Experiments on More Noise Standard Deviations}
\label{app:noiseexp}

\begin{figure}[h!]
    \centering
    \includegraphics[width=0.45\columnwidth]{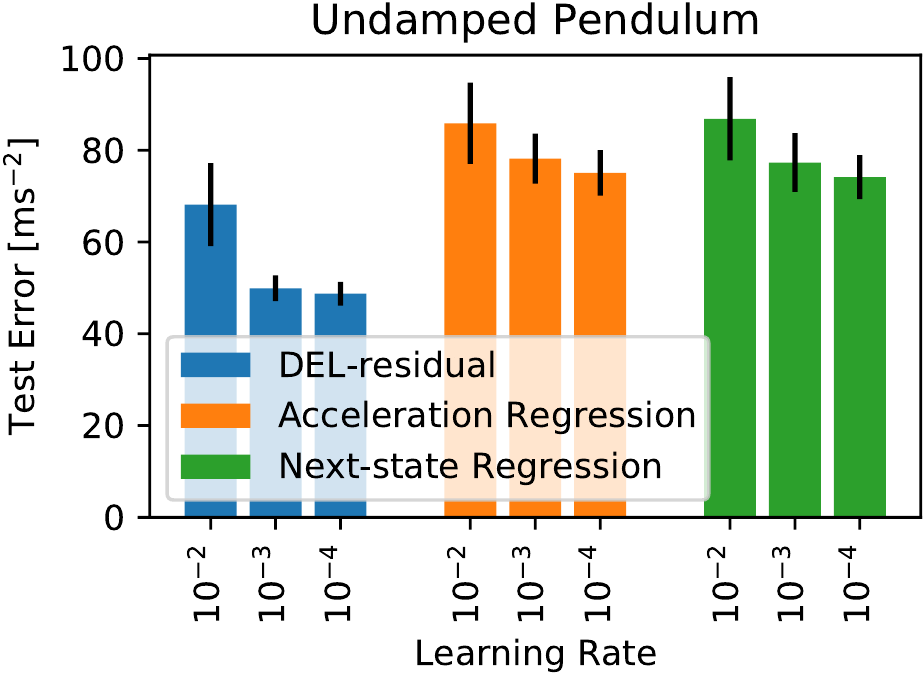}
    \includegraphics[width=0.465\columnwidth]{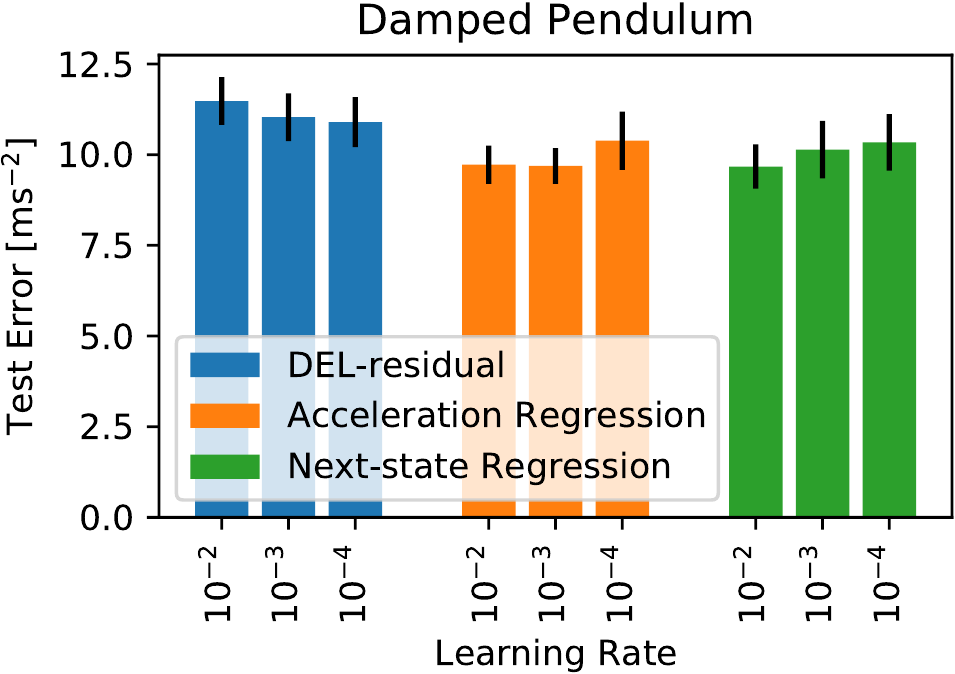}
    \caption{Comparison of training methodologies on data from a double pendulum simulated with a noise standard deviation of 0.05 $\si\radian$.}
    \label{fig:undamped005}
\end{figure}

\begin{figure}
    \centering
    \includegraphics[width=0.45\columnwidth]{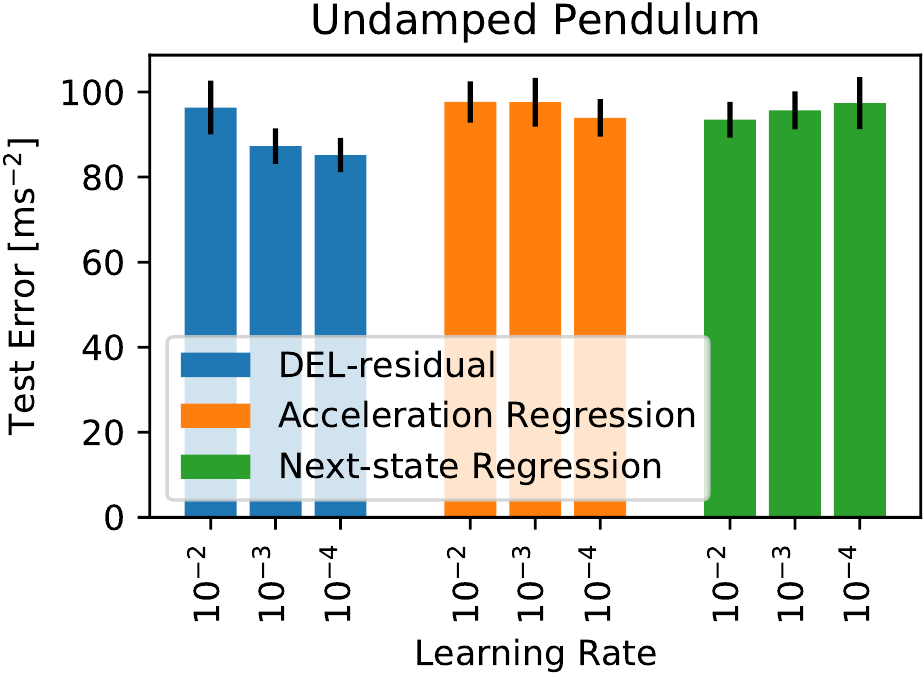}
    \includegraphics[width=0.45\columnwidth]{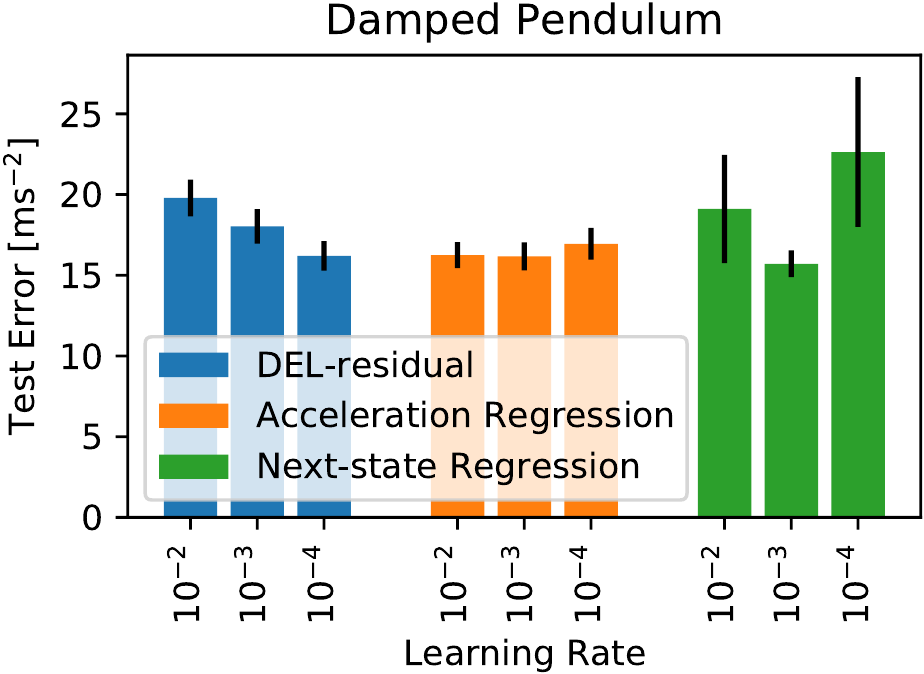}
    \caption{Comparison of training methodologies on data from a double pendulum simulated with a noise standard deviation of 0.4 $\si\radian$.}
    \label{fig:undamped04}
\end{figure}

In \Cref{fig:undamped005} and \Cref{fig:undamped04}, we compare methodologies on data from undamped and damped pendulums simulated with noise standard deviations of 0.05 and 0.4 $\si\radian$ respectively.

As we can see, training with our methodology continues to outperform acceleration and next-state regression on the undamped pendulum, though has comparable performance on a damped pendulum. 
We suspect that the reason for this difference is the fact that as the damped signal decays, noise dominates the null signal, thereby equalizing the performance of methodologies.

\end{document}